\documentclass[preprint,12pt]{elsarticle}

\usepackage{amsmath}
\usepackage{amssymb}

\usepackage{graphicx}

\usepackage{algorithm}
\usepackage{algpseudocode}

\usepackage{booktabs}    
\usepackage{multirow}
\usepackage{adjustbox}

\usepackage{siunitx}
\usepackage{microtype}

\journal{Geoenergy Science and Engineering}

\begin{document}

\begin{frontmatter}



\title{WLFM: A Well-Logs Foundation Model for Multi-Task and Cross-Well Geological Interpretation}


\author[label1]{Zhenyu Qi}
\author[label1]{Qing Yu}
\author[label2]{Jichen Wang}
\author[label2]{Yun-Bo Zhao\corref{cor1}}
\author[label3]{Zerui Li\corref{cor2}}
\author[label2]{Wenjun Lv}

\address[label1]{Institute of Advanced Technology, University of Science and Technology of China, Hefei, 230088, China}
\address[label2]{Department of Automation, University of Science and Technology of China, Hefei, 230026, China}
\address[label3]{Institute of Artificial Intelligence, Hefei Comprehensive National Science Center, Hefei, 230088, China}

\cortext[cor1]{Corresponding author: Yun-Bo Zhao (e-mail: ybzhao@ustc.edu.cn)}
\cortext[cor2]{Corresponding author: Zerui Li (e-mail: lzerui@ustc.edu.cn)}

\begin{abstract}
Well-log interpretation is fundamental for subsurface characterization but remains challenged by heterogeneous tool responses, noisy signals, and limited labels. We propose WLFM, a foundation model pretrained on multi-curve logs from 1200 wells, comprising three stages: tokenization of log patches into geological tokens, self-supervised pretraining with masked-token modeling and stratigraphy-aware contrastive learning, and multi-task adaptation with few-shot fine-tuning. WLFM consistently outperforms state-of-the-art baselines, achieving 0.0041 MSE in porosity estimation and 74.13\% accuracy in lithology classification, while WLFM–Finetune further improves to 0.0038 MSE and 78.10\% accuracy. Beyond predictive accuracy, WLFM exhibits emergent layer-awareness, learns a reusable geological vocabulary, and reconstructs masked curves with reasonable fidelity, though systematic offsets are observed in shallow and ultra-deep intervals. Although boundary detection is not explicitly evaluated here, clustering analyses suggest strong potential for future extension. These results establish WLFM as a scalable, interpretable, and transferable backbone for geological AI, with implications for multi-modal integration of logs, seismic, and textual data.
\end{abstract}



\begin{keyword}
Well-logs \sep Foundation models \sep Geological interpretation
\end{keyword}

\end{frontmatter}



\section{Introduction}
\label{sec:intro}

\begin{figure}[!h]
    \centering
    \includegraphics[width=\linewidth]{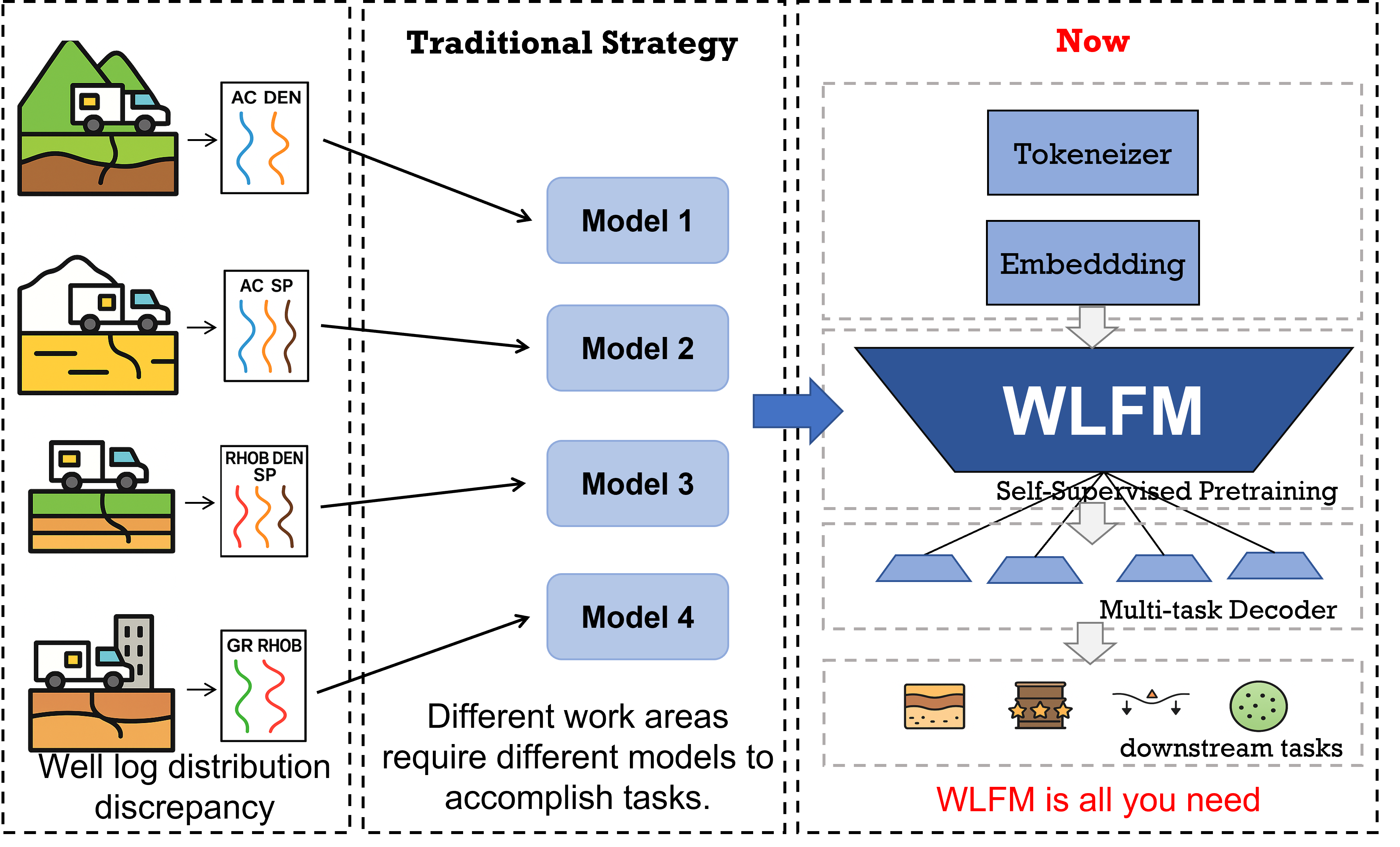}
    \caption{Conceptual motivation of the proposed WLFM framework. Well-log interpretation is challenged by noisy, incomplete, and heterogeneous multi-curve data. Traditional rule-based or task-specific models lack scalability and cross-domain generalization. WLFM introduces a tokenization-based foundation modeling approach that bridges domain-informed priors with large-scale representation learning.}
    \label{fig:WLFM_motivation}
\end{figure}

Well logging provides high-resolution measurements of subsurface formations and plays a central role in lithology classification, boundary detection, and reservoir characterization \cite{Guoyin2015Dolomite,Zhu2021NMR,Xu2022NMR}. It underpins critical applications including unconventional reservoir evaluation, carbon storage monitoring, and basin-scale geological modeling. However, practical interpretation is often constrained by noisy signals, missing curves, and heterogeneous data distributions across wells, which substantially limit the scalability and generalization of conventional workflows \cite{Gama2025ImputationBenchmark,AlFakih2025GANImputation,Antariksa2023ImputationDL}.

\subsection{Limitations of Existing Machine Learning Approaches}

The rise of machine learning has created new opportunities for automating and scaling well-log interpretation. Compared with rule-based approaches, neural networks can process large datasets, capture nonlinear dependencies, and improve predictive accuracy across diverse geological settings \cite{Xu2021SandstoneCNN,Yu2023Porosity,Chang2021SegLog,Lv2023LogRegX}. Early studies employing convolutional neural networks (CNNs) and recurrent neural networks (RNNs) demonstrated promise in lithology identification \cite{Wang2019CNN,Yang2021ML,Shi2024LithofaciesPrediction,Khan2024DNNLithofacies,Qu2025RNNLithology}, porosity prediction \cite{Qi2025CNNBiLSTMTransformer,Cheng2025CNNTransformerDensity}, and thin-bed facies analysis \cite{Jin2023ThinBedGRSL}. Yet, these approaches typically required extensive labeled data and exhibited poor transferability across formations. Domain adaptation strategies, such as maximum mean discrepancy (MMD) \cite{Chang2021MMD} and partial adaptation \cite{Li2024PDA}, attempted to mitigate cross-well distribution gaps, but relied heavily on heuristic assumptions and showed limited robustness under sparse supervision. More recent adversarial and GAN-based methods \cite{Sun2023Energies,AlFakih2025GAN,Qian2024CNNBiLSTM} enhanced transferability but introduced architectural complexity and remained vulnerable to noise and missing data.

\subsection{Toward Foundation Models in Geoscience}

Beyond well-log analysis, deep learning has demonstrated broader potential in geophysical interpretation. For instance, Tang \emph{et al.}~\cite{Tang2024Forecast} applied neural networks to transient electromagnetic time-series for subsurface body prediction, highlighting the value of sequence representation learning in geoscience. In seismic studies, foundation-style architectures have been applied to inversion \cite{Zhang2024GeoInversionTransformer,Zhang2023PretrainInversion}, seismic fault segmentation \cite{Cao2024SwinUNETR}, and cross-well prediction \cite{Lv2024GeoFormer}. Hyperspectral and geophysical time-series analysis have similarly benefited from foundation-style encoders \cite{Liu2024FactoFormer,Qi2024GRSLFoundation}. These advances collectively signal a paradigm shift from task-specific models toward domain-informed foundation models in Earth sciences.

Meanwhile, foundation models have revolutionized natural language processing and computer vision by leveraging large-scale pretraining and token-based representation learning. Large language models (LLMs) employ subword tokenization (e.g., byte pair encoding, SentencePiece), while vision and seismic foundation models tokenize images or traces into patch embeddings or vector-quantized (VQ) tokens. These designs have proven effective for scaling Transformer architectures, stabilizing masked modeling objectives, and enabling cross-modal alignment. In geophysics, Chen \emph{et al.}~\cite{Chen2025Gaia} proposed \emph{Gaia}, a pretrained model for logging-curve reconstruction. Other studies explored generic time-series foundation models such as TOTEM \cite{Talukder2025TOTEM}, TimeGPT \cite{Koeshidayatullah2024TimeSeriesFoundation,Garza2023TimeGPT1}, and conformal prediction frameworks \cite{Achour2025FoundationModelsCP}. Broader reviews also highlight both the opportunities and challenges of subsurface foundation models \cite{Koeshidayatullah2024PetroAI}. However, most existing approaches either target single-curve reconstruction or remain domain-agnostic, leaving unresolved the compounded challenges of multi-curve heterogeneity, stratigraphy-aware representation, and multi-task geological interpretation.

\subsection{This Work: WLFM}

Our key insight is to extend the tokenization paradigm to well-log interpretation. Instead of encoding continuous log patches directly as floating-point embeddings, we introduce a vector-quantized (VQ) tokenizer that discretizes multi-curve inputs into a symbolic space. This places WLFM within the same modeling framework as language and seismic foundation models, where tokens serve as the fundamental unit of representation. The unified token space offers three principal advantages: (i) a natural interface for multi-modal alignment, enabling well logs, seismic, and geological text to be jointly modeled in a shared embedding space; (ii) a structured and interpretable ``geological vocabulary,'' where tokens capture recurring morphologies such as shale–sand transitions or high-resistivity, low-porosity intervals; and (iii) improved robustness and generalization, as discrete tokens suppress amplitude shifts and noise while providing stable targets for masked modeling.

Building on this principle, we propose WLFM, one of the first foundation models explicitly designed for multi-curve well-log interpretation. WLFM is pretrained from scratch on large-scale unlabeled datasets using a modular framework that combines tokenization, self-supervised representation learning, and multi-task fine-tuning. It incorporates domain-informed priors—relative-depth encodings to preserve stratigraphic context and curve-type embeddings to represent geophysical modalities—and employs a hardware-aware pretraining pipeline to ensure efficient large-scale training.

\subsection{Contributions}

The contributions of this work are threefold:
\begin{itemize}
    \item We propose WLFM, a domain-specialized foundation model for multi-curve well-log interpretation, introducing a VQ tokenizer that transforms raw signals into a reusable ``geological vocabulary'' of discrete tokens.
    \item We design a structurally grounded pretraining framework that integrates stratigraphy-aware contrastive learning with domain-informed priors and a hardware-aware patch loading strategy for efficient training at scale.
    \item We demonstrate that WLFM, following a pretrain–then–finetune paradigm, consistently outperforms supervised and self-supervised baselines across lithology, porosity, reconstruction, and boundary tasks—particularly under cross-well, noisy, and low-label regimes. Furthermore, WLFM produces interpretable stratigraphic representations and achieves improved training efficiency.
\end{itemize}

\section{Methods}
\label{sec:methods}

\begin{figure*}[htbp]
    \centering
    \includegraphics[width=\linewidth]{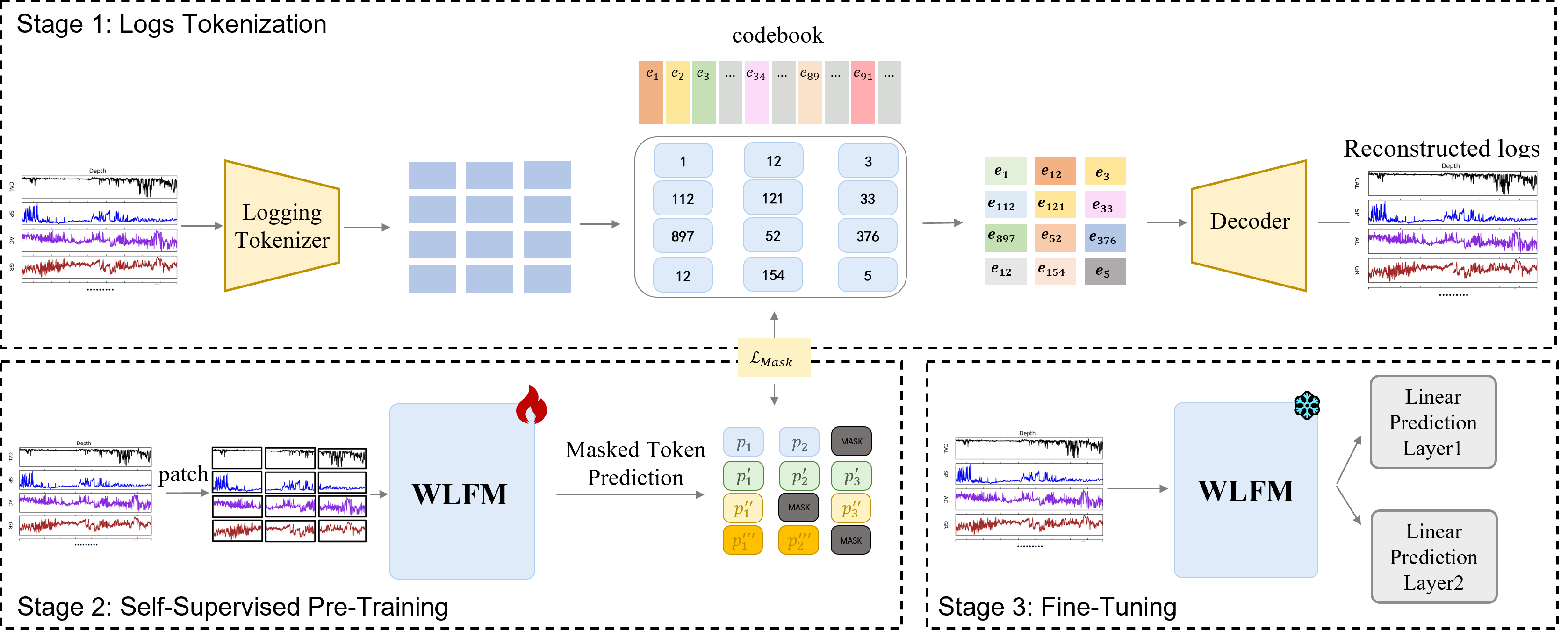}
    \caption{\textbf{Detailed architecture of WLFM.} The workflow comprises three stages: 
    (1) \emph{Tokenization}—depth-aligned log patches are discretized into geological tokens via vector quantization; 
    (2) \emph{Self-supervised pretraining}—a transformer backbone is optimized with masked-token modeling and stratigraphy-aware contrastive learning to capture both intra-log structure and cross-well semantics; 
    (3) \emph{Task adaptation}—lightweight heads are fine-tuned for lithology classification, porosity regression, and curve reconstruction.}
    \label{fig:WLFM_architecture}
\end{figure*}

We propose WLFM, a modular three-stage framework for well-log representation learning (Fig.~\ref{fig:WLFM_architecture}). 
Stage~1 converts continuous multi-curve logs into discrete tokens via a domain-aware tokenizer. 
Stage~2 pretrains a transformer backbone using masked-token modeling and stratigraphy-aware contrastive learning. 
Stage~3 enables downstream adaptation via multi-task fine-tuning. 
This design allows WLFM to learn from heterogeneous formations and generalize under limited supervision.

\subsection{Domain-Aware Encoder for Multi-Curve Well Logs}
Well logging records consist of heterogeneous physical measurements acquired at uniform depth intervals. 
These measurements—such as gamma ray (GR), spontaneous potential (SP), and acoustic travel-time (AC)—have distinct statistics and reflect different geological properties. 
Furthermore, the vertical organization of well-log sequences encodes stratigraphic information essential for tasks like lithology classification and porosity estimation \cite{Shi2024LithofaciesPrediction, Gama2025ImputationBenchmark, AlFakih2025GANImputation, Antariksa2023ImputationDL, Khan2024DNNLithofacies, Qu2025RNNLithology}. 
However, existing deep learning and GAN-based methods often generalize poorly across wells, show limited robustness to geological variability, and lack explicit stratigraphic awareness.

To address these issues, our encoder (Stage~1 of Fig.~\ref{fig:WLFM_architecture}) embeds both modality-specific characteristics and stratigraphic context into the learned representations. 
Given an input patch \(x \in \mathbb{R}^{C \times L}\) (channels \(C\), depth window \(L\)), we enrich the raw data with two priors: 
(i) a trainable \emph{curve-type} embedding per channel (e.g., GR, SP, AC), enabling modality-aware attention; and 
(ii) \emph{relative-depth} positional encodings to preserve stratigraphic order. 
The combined representation is processed by a lightweight convolutional module (depthwise separable convolutions with residual connections), capturing local morphologies while remaining scalable for large-scale pretraining. 
The encoder outputs latent features \(z_e\) for tokenization and representation learning.

\subsection{Patch-wise Representation and Tokenization}
As illustrated in Stage~1 of Fig.~\ref{fig:WLFM_architecture}, we transform encoded patches into a compact, discrete token sequence that abstracts recurring geological patterns. 
Rather than modeling raw multi-curve inputs directly, we employ vector quantization (VQ) to discretize \(z_e \in \mathbb{R}^{d}\) into a learned codebook \(\mathcal{E}=\{e_k\}_{k=1}^K\):
\begin{equation}
z_q = \arg\min_{e_k \in \mathcal{E}} \big\| z_e - e_k \big\|_2 .
\end{equation}
We jointly optimize encoder, decoder, and codebook with a VQ-VAE loss:
\begin{equation}
\mathcal{L}_{\text{token}} = \big\| x - \hat{x} \big\|_2^2 + \big\| \text{sg}[z_e] - e_k \big\|_2^2 + \beta \big\| z_e - \text{sg}[e_k] \big\|_2^2 ,
\end{equation}
where \(\hat{x}\) is the reconstruction, \(\text{sg}[\cdot]\) denotes stop-gradient, and \(\beta\) controls codebook commitment. 
The resulting token sequence forms a symbolic abstraction of the subsurface—analogous to words in language—serving as input to pretraining.

\paragraph{Rationale for Discretization.}
Directly training on continuous patches tends to encode amplitude offsets, tool biases, and noise as semantics, hindering cross-well alignment and destabilizing masked \emph{regression}. 
Mapping recurring multi-curve morphologies to a finite codebook yields: 
(i) well-invariant, noise-robust abstractions for stratigraphy-aware contrast; 
(ii) a stable masked-token \emph{classification} target with clearer gradients; and 
(iii) improved label efficiency with graceful degradation when some curves are missing.

\paragraph{Tokenizer Implementation.}
Patches (length \(L\), stride \(s\)) are z-scored per well; RT can be log-transformed. 
Curve-type and relative-depth embeddings are added \emph{before} quantization. 
The codebook size \(K\), embedding dimension \(d\), commitment \(\beta\), and EMA updates are used; dead codes are periodically re-initialized from high-loss samples. 
Exact \((K,d,\beta,L,s)\) are reported in the experimental setup.

\subsection{Self-Supervised Pretraining}
Stage~2 in Fig.~\ref{fig:WLFM_architecture} pretrains WLFM with a dual-objective strategy to capture local continuity within wells and stratigraphic consistency across wells.

\subsubsection{Masked Token Modeling (MTM)}
Given a set of masked positions \(\mathcal{M}\) with ground-truth token indices \(p_i\) and predictions \(\hat{p}_i\), we minimize
\begin{equation}
\mathcal{L}_{\text{MTM}} = \sum_{i \in \mathcal{M}} \text{CE}(\hat{p}_i, p_i).
\end{equation}
Block-wise masking along depth with ratio \(r\) is adopted, which encourages the model to recover stratigraphic context rather than isolated tokens.

\subsubsection{Stratigraphy-Aware Contrastive Learning (SCL)}
For a token embedding \(z_i\) and its stratigraphic counterpart \(z_j^+\) from another well, with in-batch negatives \(\{z_k\}\), we apply an InfoNCE loss:
\begin{equation}
\mathcal{L}_{\text{SCL}} = -\log \frac{\exp(\text{sim}(z_i, z_j^+)/\tau)}{\sum_{k} \exp(\text{sim}(z_i, z_k)/\tau)},
\end{equation}
where \(\text{sim}(\cdot,\cdot)\) denotes cosine similarity and \(\tau\) is the temperature hyperparameter.

\subsubsection{Positive-Pair Construction}
Cross-well positives are constructed via relative-depth alignment within a tolerance and further filtered by low-frequency similarity (Pearson correlation \(>\tau_{\text{sim}}\) after band-pass). 
We also examine two variants: (i) a purely depth-based matcher without similarity filtering, and (ii) an anchored matcher that incorporates sparse human-labeled layer tops propagated to local neighborhoods. 
Negatives include distant segments within the same well and non-aligned cross-well windows. 
In practice, stratigraphic anchors are available only for a limited subset of wells, 
so the SCL term acts as a weak regularizer while the majority of wells are trained with MTM alone.

\subsubsection{Total Loss}
The overall pretraining objective combines both terms:
\begin{equation}
\mathcal{L}_{\text{pretrain}} = \mathcal{L}_{\text{MTM}} + \alpha \,\mathcal{L}_{\text{SCL}}
\end{equation}
where \(\alpha\) balances the contribution of the contrastive regularization. 
Hyperparameters \((r,\tau,\tau_{\text{sim}},\alpha)\) are provided in the experimental setup.

\subsection{Multi-Task Fine-Tuning}
\label{subsec:finetune}
Finally, \textbf{Stage~3} in Fig.~\ref{fig:WLFM_architecture} attaches lightweight heads to the pretrained encoder for lithology classification, porosity regression, and curve reconstruction. 
Unless specified, the encoder is frozen and only task heads are optimized. 
This stage corresponds to WLFM–Finetune, where a small number of labeled wells are used for few-shot adaptation under a multi-task setting.

\paragraph{Objectives.}
Let \(\mathcal{M}\) be the reconstruction mask set. 
The overall loss is
\begin{equation}
\mathcal{L}_{\text{multi}}=\lambda_r\mathcal{L}_{\text{recon}}+\lambda_p\mathcal{L}_{\text{poro}}+\lambda_l\mathcal{L}_{\text{litho}}+\gamma\mathcal{L}_{\text{consistency}},
\end{equation}

with
\begin{align}
\mathcal{L}_{\text{recon}} &= \frac{1}{|\mathcal{M}|}\sum_{(c,t)\in\mathcal{M}}\big\|\hat{x}_{c,t}-x_{c,t}\big\|_2^2, \\
\mathcal{L}_{\text{poro}}  &= \frac{1}{T}\sum_t \big|\hat{y}^{\text{poro}}_t-y^{\text{poro}}_t\big|, \\
\mathcal{L}_{\text{litho}} &= -\sum_{t,k} y^{\text{litho}}_{t,k}\log \hat{y}^{\text{litho}}_{t,k}, \\
\mathcal{L}_{\text{consistency}} &= \mathbb{E}_t\!\left[\mathrm{KL}\big(P_{\text{poro}}\parallel P_{\text{litho}}\big)\right],
\end{align}
where the KL term encourages physical plausibility (e.g., higher porosity consistent with sandstones).

\paragraph{Modality Dropout.}
To improve robustness to missing curves, we simulate missing modalities during pretraining and fine-tuning by randomly dropping channel groups with probability \(p\) drawn from the test-time distribution.

\subsection{Hardware-Aware Pretraining Pipeline}
\begin{algorithm}[htbp]
\caption{Hardware-Aware Patch Loader for Pretraining}
\label{alg:hw_aware_loader}
\begin{algorithmic}[1]
\Require Dataset \(\mathcal{D}\), Encoder \(\mathcal{E}\), Score threshold \(\tau\)
\While{not converged}
    \State Initialize shared queue \(\mathcal{Q}\)
    \ForAll{CPU workers in parallel}
        \State Sample well \(\mathcal{W} \sim \mathcal{D}\), extract patches \(p\)
        \For{each patch \(p\)}
            \State Compute \(\text{Score}(p)\)
            \If{\(\text{Score}(p) > \tau\)}
                \State Encode \(p \rightarrow z\), enqueue \(z\) into \(\mathcal{Q}\)
            \EndIf
        \EndFor
    \EndFor
    \State GPU dequeues tokens from \(\mathcal{Q}\), updates WLFM
\EndWhile
\end{algorithmic}
\end{algorithm}

To maximize throughput on large datasets, we select high-value patches using stratigraphic and statistical cues and feed them asynchronously. 
A depth-aware term highlights layer boundaries via vertical gradients; a channel-aware term favors high inter-log variance:
\begin{equation}
\text{Score}(p) = \lambda_1 \cdot \nabla_{\text{depth}}(p) + \lambda_2 \cdot \text{Var}_{\text{channel}}(p).
\end{equation}
Low-frequency summaries for similarity filtering and per-patch scores are computed on CPUs and cached; only selected, tokenized patches are sent to GPUs. 
On an NVIDIA RTX A6000 (batch size 256, eight workers, AMP), throughput increases from \(275 \pm 6\) to \(680 \pm 8\) patches/s (\(\sim 2.5\times\)) and average GPU utilization rises from 43\% to 91\%.

\paragraph{Reproducibility Protocol.}
All splits are at the \emph{well} level to avoid leakage. 
We report mean$\pm$std over five seeds and apply early stopping on validation. 
Key hyperparameters \((K,d,\beta,L,s,r,\tau,\tau_{\text{sim}},\alpha,p)\), optimizer settings, and hardware are detailed in the experimental setup. 
Statistical testing uses two-sided paired $t$-tests versus baselines.

\begin{table}[htbp]
  \caption{Porosity and Lithology Performance (\textit{mean ± std} over 5 seeds; best in bold).}
  \label{tab:overall_perf_final}
  \centering
  \small
  \renewcommand{\arraystretch}{1.12}
  \setlength{\tabcolsep}{3.5pt}
  \begin{adjustbox}{max width=\linewidth}
  \begin{tabular}{
    @{} l
    S S  
    S S  
    S S  
    S    
    @{}}
    \toprule
    & \multicolumn{6}{c}{\textbf{Porosity (lower is better)}} & \multicolumn{1}{c}{\textbf{Lithology Acc. / \%}} \\
    \cmidrule(lr){2-7}
    & \multicolumn{2}{c}{\textbf{Dataset A}} & \multicolumn{2}{c}{\textbf{Dataset B}} & \multicolumn{2}{c}{\textbf{Dataset C}} & \textbf{Dataset C} \\
    \cmidrule(lr){2-3}\cmidrule(lr){4-5}\cmidrule(lr){6-7}
    \textbf{Method} & {MAE} & {MSE} & {MAE} & {MSE} & {MAE} & {MSE} & {Acc} \\
    \midrule
    CNN            & \num{0.0438 \pm 0.0023} & \num{0.0087 \pm 0.0004} & \num{0.0455 \pm 0.0034} & \num{0.0073 \pm 0.0004} & \num{0.0436 \pm 0.0021} & \num{0.0053 \pm 0.0002} & \num{68.95 \pm 2.35} \\
    LSTM           & \num{0.0414 \pm 0.0032} & \num{0.0059 \pm 0.0007} & \num{0.0664 \pm 0.0035} & \num{0.0059 \pm 0.0007} & \num{0.0425 \pm 0.0019} & \num{0.0050 \pm 0.0002} & \num{69.10 \pm 1.34} \\
    CNN--LSTM$^{\ast}$ & \num{0.0406 \pm 0.0022} & \num{0.0057 \pm 0.0026} & \num{0.0531 \pm 0.0020} & \num{0.0056 \pm 0.0005} & \num{0.0425 \pm 0.0013} & \num{0.0050 \pm 0.0002} & \num{69.85 \pm 4.30} \\
    Transformer    & \num{0.0407 \pm 0.0012} & \num{0.0057 \pm 0.0006} & \num{0.0442 \pm 0.0013} & \num{0.0055 \pm 0.0002} & \num{0.0405 \pm 0.0009} & \num{0.0051 \pm 0.0002} & \num{65.01 \pm 1.42} \\
    \midrule
    WLFM--400      & \num{0.0395 \pm 0.0011} & \num{0.0055 \pm 0.0004} & \num{0.0429 \pm 0.0015} & \num{0.0049 \pm 0.0005} & \num{0.0396 \pm 0.0011} & \num{0.0039 \pm 0.0001} & \num{70.10 \pm 1.28} \\
    WLFM--600      & \num{0.0388 \pm 0.0010} & \num{0.0052 \pm 0.0006} & \num{0.0416 \pm 0.0013} & \num{0.0052 \pm 0.0006} & \num{0.0376 \pm 0.0010} & \num{0.0039 \pm 0.0009} & \num{73.02 \pm 1.22} \\
    {\bfseries WLFM--1200} & {\bfseries \num{0.0371 \pm 0.0008}} & {\bfseries \num{0.0051 \pm 0.0005}} & {\bfseries \num{0.0391 \pm 0.0024}} & {\bfseries \num{0.0049 \pm 0.0012}} & {\bfseries \num{0.0378 \pm 0.0006}} & {\bfseries \num{0.0041 \pm 0.0008}} & {\bfseries \num{74.13 \pm 4.30}} \\
    \midrule
    {\bfseries WLFM--Finetune} & {\bfseries \num{0.0352 \pm 0.0013}} & {\bfseries \num{0.0048 \pm 0.0015}} & {\bfseries \num{0.0391 \pm 0.0026}} & {\bfseries \num{0.0049 \pm 0.0012}} & {\bfseries \num{0.0322 \pm 0.0012}} & {\bfseries \num{0.0038 \pm 0.0024}} & {\bfseries \num{78.10 \pm 0.83}} \\
    \bottomrule
  \end{tabular}
  \end{adjustbox}
  \vspace{2mm}
  \footnotesize
  $^{\ast}$Baseline used for relative percentage calculation (best non-WLFM row). All results are averaged over 5 random seeds; standard deviations shown after “$\pm$”.
\end{table}

\section{Results}
\label{sec:results}

We organize our empirical study into five dimensions: overall performance, scalability and label efficiency, latent representation and geological semantics, component contributions, and failure analysis.

\subsection{Overall Performance Benchmark}

Table~\ref{tab:overall_perf_final} summarizes the performance of WLFM and baseline models across three porosity regression datasets and one lithology classification benchmark. WLFM-1200 achieves the best results across all tasks. 

For porosity estimation on Dataset C, WLFM-1200 attains a mean squared error (MSE) of 0.0041, outperforming all traditional and deep learning baselines. In lithology classification, it reaches an accuracy of 74.13\%, surpassing the strongest baseline (CNN-LSTM, 69.85\%) by 6.1\% relative improvement.

When further fine-tuned under multi-task supervision with a small number of labeled wells, WLFM–Finetune achieves the strongest overall results (Table~\ref{tab:overall_perf_final}), including a porosity MSE of 0.0038 and lithology accuracy of 78.10\%. This demonstrates that WLFM not only provides strong pretrained features but also adapts efficiently with limited supervision.

Across all configurations, WLFM is the only model that consistently ranks first in every evaluation metric. These results validate its capacity for multi-curve representation learning and set the stage for subsequent analyses of scaling, label efficiency, and representation quality.

\subsection{Scaling Behaviour and Few-Shot Adaptation}
\label{sec:scaling}

\begin{figure}[htbp]
\centering
\includegraphics[width=\linewidth]{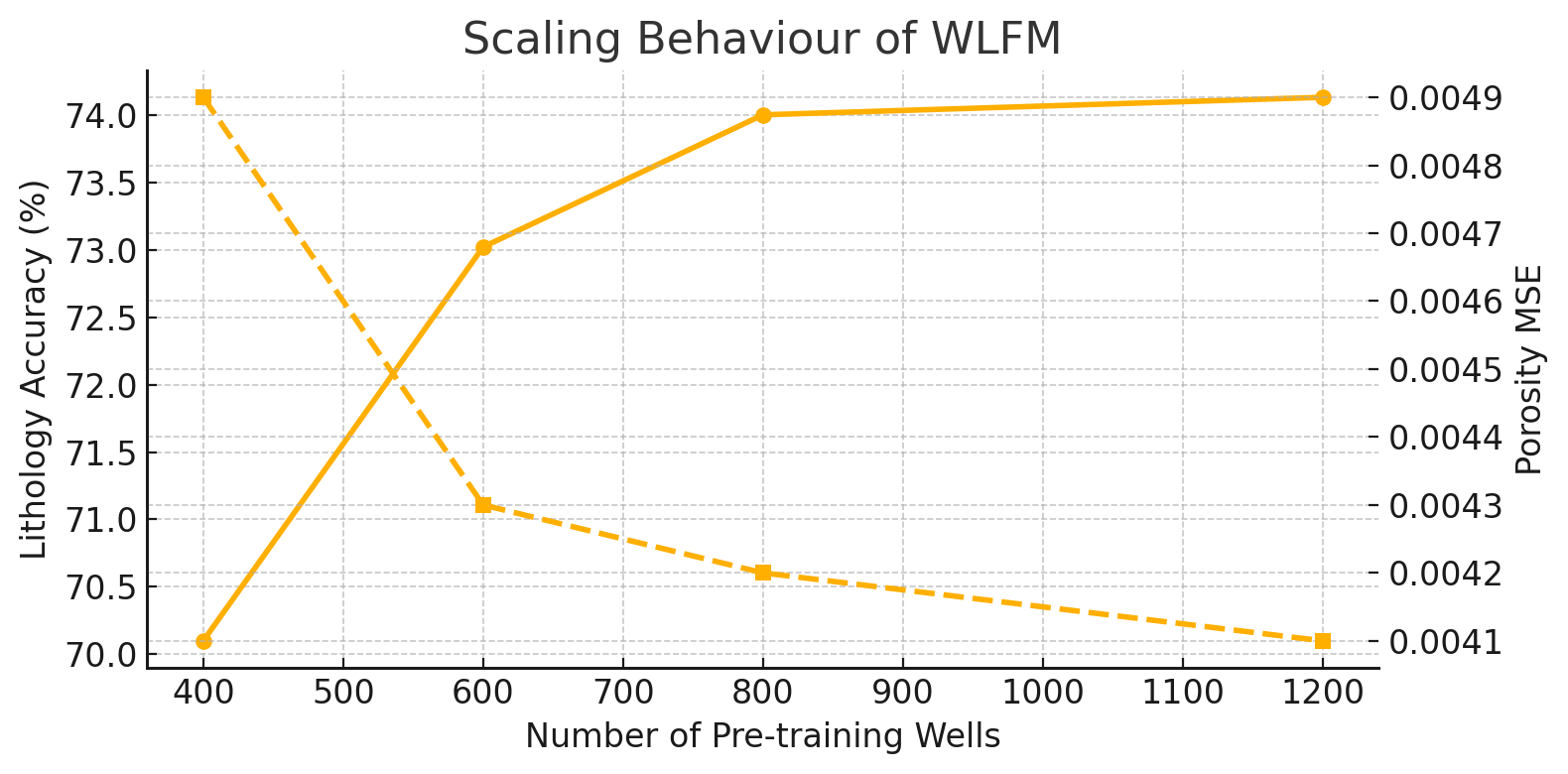}
\caption{Scaling behaviour: larger pretraining sets yield higher lithology accuracy (solid) and lower porosity MSE (dashed). Error bars show variation across three runs.}
\label{fig:scale_plot}
\end{figure}

Figure~\ref{fig:scale_plot} quantifies how WLFM benefits from larger pretraining sets. As the number of unlabeled wells increases from 400 to 1200, lithology classification accuracy improves from 70.1\% to 74.1\%, while porosity regression error (MSE) decreases from 0.0049 to 0.0041. The trend is consistent across runs and most pronounced in classification, suggesting that lithology prediction continues to benefit from larger corpora, whereas porosity estimation saturates earlier.

In addition, WLFM demonstrates strong few-shot adaptation. Table~\ref{tab:comprehensive_comparison_single_col} shows that WLFM-1200 achieves competitive porosity performance with as few as 10 labeled wells (MSE 0.0053), and further improves to 0.0047 with 30 wells. Smaller models such as WLFM-400 require more supervision to reach similar accuracy. These few-shot results are consistent with the WLFM--Finetune benchmark in Table~\ref{tab:overall_perf_final}, highlighting that multi-task fine-tuning with limited supervision unlocks the full potential of WLFM.

\begin{table}[htbp]
  \centering
  \caption{Few-shot fine-tuning on Dataset C (Porosity MSE ↓).}
  \label{tab:comprehensive_comparison_single_col}
  \renewcommand{\arraystretch}{1.15}
  \begin{tabular}{
    @{}c
    S[table-format=1.4]
    S[table-format=1.4]
    S[table-format=1.4]
    @{}}
    \toprule
    \textbf{Wells} & \textbf{WLFM-1200} & \textbf{WLFM-600} & \textbf{WLFM-400} \\ \midrule
    10 & 0.0053 & 0.0055 & 0.0061 \\
    20 & \bfseries 0.0049 & 0.0050 & 0.0050 \\
    30 & \bfseries 0.0047 & 0.0049 & 0.0049 \\ \bottomrule
  \end{tabular}
\end{table}

\subsection{Latent Representation and Geological Semantics}

\begin{figure}[!t]
\centering
\includegraphics[width=\linewidth]{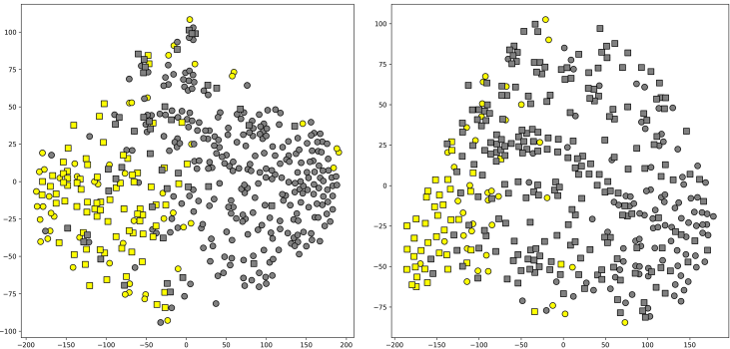}
\caption{t-SNE of token embeddings colored by lithofacies. Distinct clusters suggest interpretable latent structure.}
\label{fig:tsne_embed}
\end{figure}

Figure~\ref{fig:tsne_embed} visualizes token embeddings using t-SNE, colored by lithofacies. Despite the absence of lithology labels during pretraining, embeddings form distinct clusters that align with lithological boundaries. K-means clustering yields an Adjusted Rand Index (ARI) of 0.78 and a purity score of 82.3\%, confirming that WLFM encodes stratigraphically coherent semantics in an unsupervised setting.

\subsection{Layer-Aware Semantic Alignment}

\begin{figure*}[!t]
\centering
\includegraphics[width=\textwidth,height=0.75\textheight,keepaspectratio]{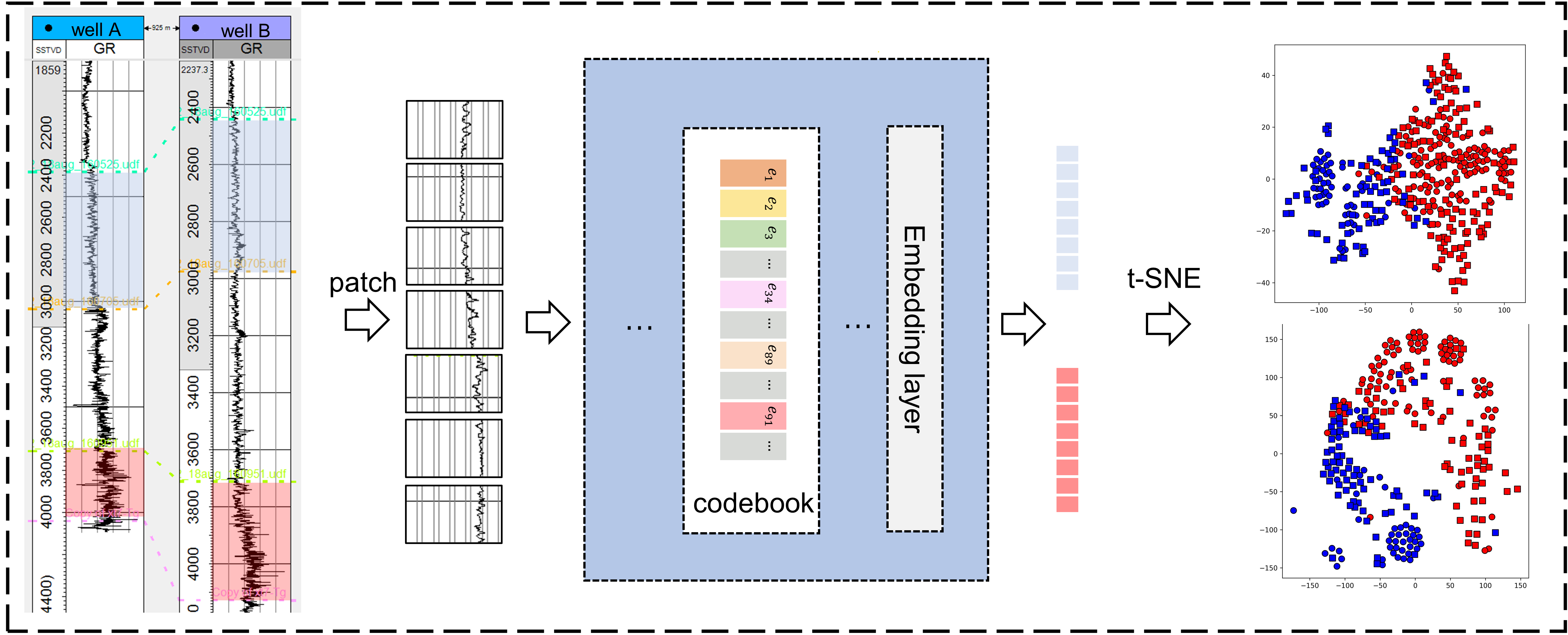}
\caption{WLFM learns layer-aware representations. Embeddings from different wells but the same stratigraphic interval form tight clusters, invariant to well identity.}
\label{fig:tsne_layer_embedding}
\end{figure*}

To evaluate domain invariance, Fig.~\ref{fig:tsne_layer_embedding} compares embeddings of patches sampled from the same stratigraphic interval across different wells. Colors denote layer identity, shapes denote well source. Embeddings corresponding to the same layer form tight clusters regardless of acquisition context, demonstrating that WLFM internalizes stratigraphic semantics and achieves cross-well alignment.

\subsection{Component Contributions and Multi-Curve Integration}
\label{subsec:ablation}

\paragraph{Tokenization and Contrastive Learning.}
Table~\ref{tab:ablation_tokenizer} shows that vector quantization and stratigraphy-aware contrastive learning are both critical. Continuous embeddings (\emph{Raw-Cont}) perform worst, VQ alone (\emph{VQ-noSCL}) improves robustness, and the full model (\emph{VQ-CE}) achieves the best results across lithology, porosity, and few-shot adaptation.

\begin{table}[!t]
\caption{Ablation: Tokenization vs. Continuous Patches on Dataset C.}
\label{tab:ablation_tokenizer}
\centering
\renewcommand{\arraystretch}{1.15}
\setlength{\tabcolsep}{5pt}
\begin{tabular}{lcccccc}
\toprule
\textbf{Method} & \textbf{Acc (\%)} & \textbf{MSE $\downarrow$} & \multicolumn{3}{c}{\textbf{Few-shot MSE $\downarrow$}} \\
\cmidrule(lr){4-6}
& & & 10 wells & 20 wells & 30 wells \\
\midrule
Raw-Cont         & 68.5 & 0.0062 & 0.0062 & 0.0058 & 0.0055 \\
VQ-noSCL         & 71.2 & 0.0052 & 0.0056 & 0.0052 & 0.0050 \\
\textbf{VQ-CE (ours)} & \textbf{74.1} & \textbf{0.0041} & \textbf{0.0053} & \textbf{0.0049} & \textbf{0.0047} \\
\bottomrule
\end{tabular}
\end{table}

\begin{figure}[htbp]
  \centering
  \includegraphics[width=\linewidth]{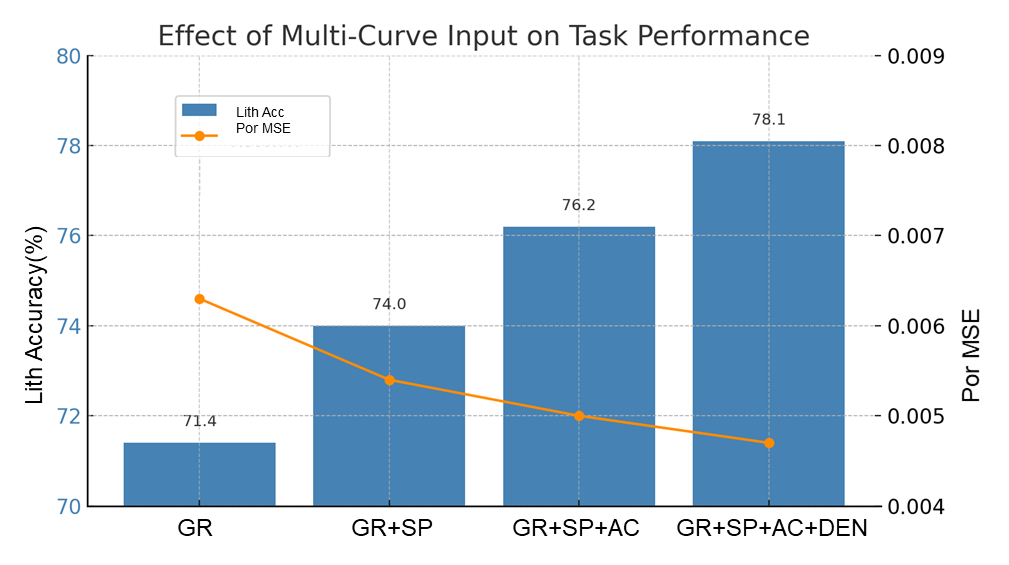}
  \caption{Influence of additional curves on lithology (Acc), porosity (MSE; lower is better). Error bars show $\pm1\sigma$ over three seeds.}
  \label{fig:multicurve}
\end{figure}

\paragraph{Multi-Curve Synergy.}
Figure~\ref{fig:multicurve} illustrates the influence of input modalities. Starting with GR alone (71.4\% lithology accuracy), adding SP, AC, and DEN progressively improves performance, culminating in 78.1\% accuracy and 0.0042 porosity MSE. This confirms that complementary curves provide substantial gains.

\paragraph{Multi-Task Supervision.}
As shown in Table~\ref{tab:multitask_compact}, multi-task training further enhances performance, reducing porosity MSE by 25\% (0.0040 $\rightarrow$ 0.0030) and moderately improving lithology accuracy (73.4\% $\rightarrow$ 75.8\%). Continuous properties benefit most, reflecting stronger cross-task synergies.

\begin{table}[htbp]
  \centering
  \caption{Effect of multi-task learning on downstream tasks (Dataset~C).}
  \label{tab:multitask_compact}
  \renewcommand{\arraystretch}{1.15}
  \begin{tabular}{
    @{}c
    S[table-format=2.2]
    S[table-format=2.2]
    S[table-format=1.4]
    @{}}
    \toprule
    {$\lambda_{\text{MT}}$} &
    \multicolumn{2}{c}{\textbf{Lithology}} &
    \multicolumn{1}{c}{\textbf{Porosity}} \\
    \cmidrule(r){2-3}
    & {Acc (\%)} & {F1} & {MSE} \\ \midrule
    0  & 73.38 & 75.90 & 0.0040 \\
    12 & \bfseries 75.78 & \bfseries 82.80 & \bfseries 0.0030 \\ 
    \bottomrule
  \end{tabular}
\end{table}

\subsection{Failure Cases and Limitations}
\label{sec:failure}

\begin{figure*}[!t]
\centering
\includegraphics[width=\textwidth,height=0.75\textheight,keepaspectratio]{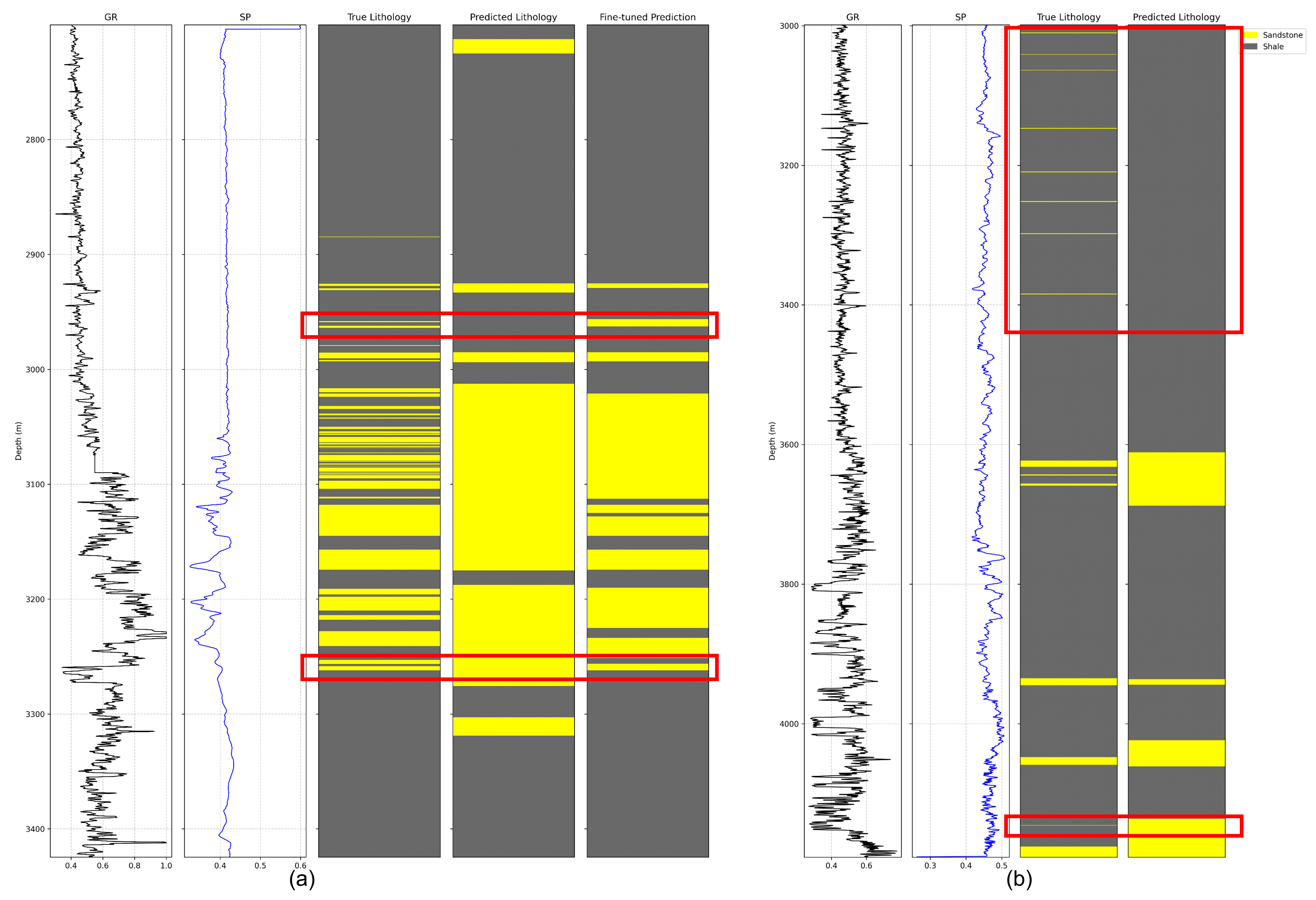}
\caption{WLFM lithology prediction under varying geological conditions. \textbf{(a)} Successful case: Fine-tuning improves token stability and enhances alignment with expert-labeled lithology, particularly within complex interbedded stratigraphy. \textbf{(b)} Failure case: In thinly layered and noisy intervals, WLFM predictions exhibit fragmented transitions and poor continuity, resulting in significant mismatch with true lithofacies boundaries.}
\label{fig:failure}
\end{figure*}

Figure~\ref{fig:failure} shows representative successes and failures. Fine-tuning improves token stability in complex interbedded zones (Fig.~\ref{fig:failure}a), but WLFM struggles with thinly interbedded and noisy formations (Fig.~\ref{fig:failure}b), where predictions fragment and continuity degrades. These cases highlight the need for hierarchical tokenization and uncertainty-aware modeling in future work.

\subsection{Summary of Results}

Overall, WLFM achieves state-of-the-art performance across porosity regression and lithology classification (Table~\ref{tab:overall_perf_final}), scales effectively with larger pretraining sets (Fig.~\ref{fig:scale_plot}), adapts efficiently under few-shot settings (Table~\ref{tab:comprehensive_comparison_single_col}), and learns interpretable latent representations (Figs.~\ref{fig:tsne_embed}, \ref{fig:tsne_layer_embedding}). Component analysis (Table~\ref{tab:ablation_tokenizer}, Fig.~\ref{fig:multicurve}, Table~\ref{tab:multitask_compact}) confirms that tokenization, multi-curve integration, and multi-task supervision jointly contribute to robust generalization. These results establish WLFM as a scalable and transferable foundation model for intelligent well-log interpretation.

\section{Discussion}
\label{sec:discussion}

This section synthesizes the main findings of our experiments, emphasizing geological interpretation, model design, and broader implications. We structure the discussion into four themes: (i) tokenization and latent semantics, (ii) scaling and generalization, (iii) task adaptation and multi-modal potential, and (iv) biases, limitations, and future directions.

\subsection{Tokenization and Latent Semantics}
WLFM’s pretraining yields embeddings that are both semantically rich and geologically meaningful. As shown in Fig.~\ref{fig:tsne_embed}, token embeddings form distinct clusters aligned with lithofacies, despite the absence of lithology supervision. Moreover, the learned codebook exhibits emergent layer-awareness (Fig.~\ref{fig:tsne_layer_embedding}), even though stratigraphic labels were only sparsely available. These results confirm that tokenization and self-supervised learning provide strong inductive bias for propagating stratigraphic semantics across unlabeled wells.

Ablation experiments (Table~\ref{tab:ablation_tokenizer}) further demonstrate that tokenization is indispensable. Continuous embeddings (\emph{Raw-Cont}) overfit to local signals and degrade under cross-well transfer, while tokenization without contrastive learning (\emph{VQ-noSCL}) improves robustness but lacks stratigraphic consistency. The full model (\emph{VQ-CE}) achieves the best lithology accuracy, porosity regression error, and few-shot adaptation, establishing tokenization as a prerequisite for structured abstraction and robust generalization.

\subsection{Scaling Behaviour and Generalization}
Scaling experiments (Fig.~\ref{fig:scale_plot}) reveal that larger pretraining sets steadily improve downstream performance: lithology accuracy increases from 70.1\% (WLFM-400) to 74.1\% (WLFM-1200), while porosity MSE decreases from 0.0049 to 0.0041. Few-shot results (Table~\ref{tab:comprehensive_comparison_single_col}) show that WLFM-1200 reaches 0.0047 MSE with only 10 labeled wells, whereas smaller models require far more supervision. These findings highlight that diversity of unlabeled wells—spanning lithologies, basins, and acquisition conditions—is more valuable than additional labels, underscoring the need for global-scale pretraining repositories to maximize transferability.

\subsection{Task Adaptation and Multi-Modal Potential}
WLFM demonstrates strong task adaptability but with task-specific benefits. Multi-task supervision (Table~\ref{tab:multitask_compact}) substantially improves porosity estimation (MSE reduced by 25\%), while lithology classification shows smaller gains due to the discrete and often ambiguous nature of facies boundaries. Input ablations (Fig.~\ref{fig:multicurve}) confirm that incorporating multiple log curves boosts both lithology and porosity performance, with acoustic and spontaneous potential logs contributing the largest gains.

Beyond single-modality logs, tokenization enables seamless integration with other modalities. Since seismic foundation models, large language models, and WLFM all operate on tokenized inputs, they can be aligned within a shared symbolic space. This facilitates joint modeling of logs, seismic traces, and geological text, supports interpretable mappings between modalities, and provides scalable storage and retrieval mechanisms. Thus, WLFM not only advances well-log interpretation but also lays the foundation for multi-modal geoscience AI.

\subsection{Biases, Limitations, and Future Directions}

\begin{figure}[!t] \centering \includegraphics[width=\linewidth]{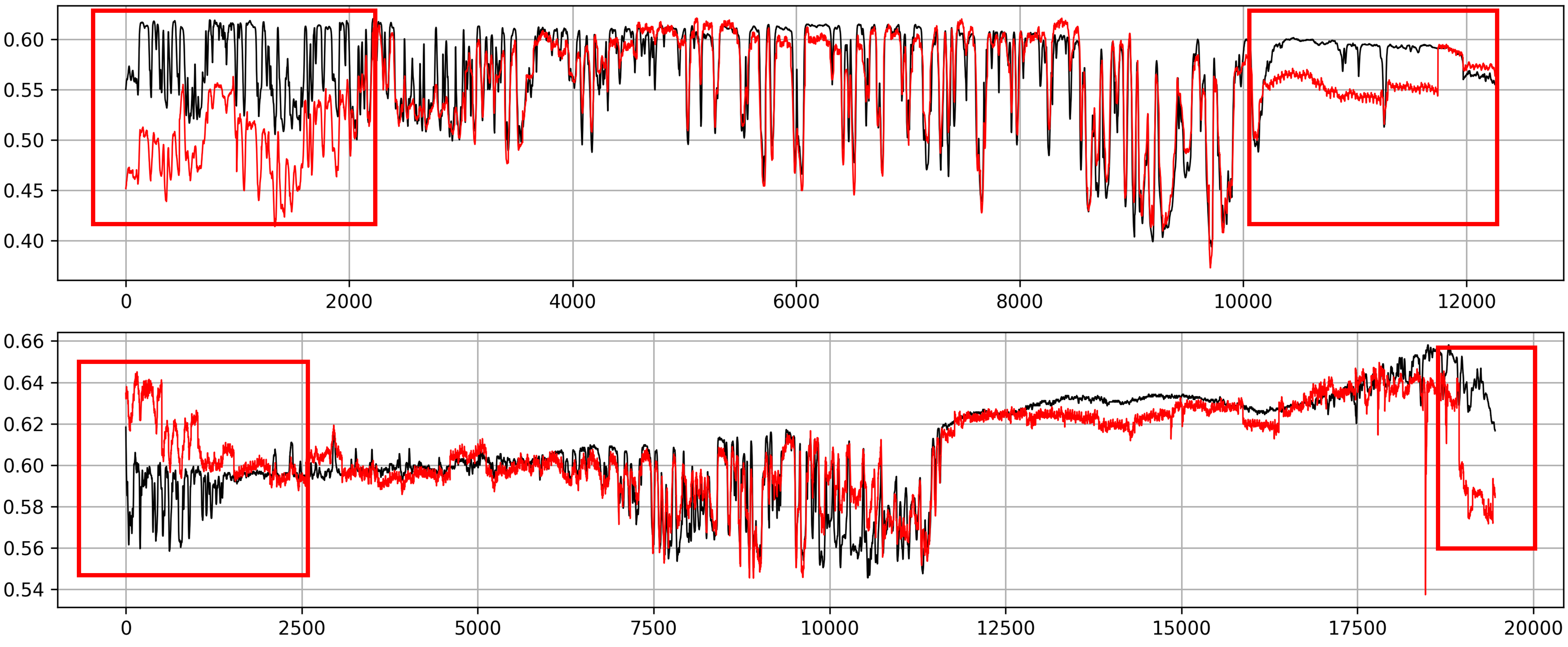} \caption{Examples of curve reconstruction failures. While WLFM successfully reconstructs the shape of masked log segments, systematic value offsets are observed in the shallow (top) and ultra-deep (bottom) intervals. Red curves: WLFM predictions; black curves: ground truth logs. These discrepancies often align with borehole transitions such as casing boundaries or sensor recalibration zones.} \label{fig:offset} \end{figure}

While WLFM shows strong robustness, certain operational and geological challenges remain. As illustrated in Fig.~\ref{fig:offset}, systematic value offsets appear in shallow and ultra-deep intervals, likely reflecting casing transitions or sensor recalibration rather than geology. Though not detrimental to reservoir-scale interpretation, these sensitivities may provide auxiliary value for engineering tasks such as anomaly or casing detection. Failure cases (Fig.~\ref{fig:failure}) highlight difficulties in noisy and thinly interbedded formations, where predictions fragment and lithology continuity degrades. Addressing such cases may require finer-resolution tokenization, hierarchical encoders, or uncertainty-aware objectives. Furthermore, scaling to millions of wells will demand distributed pretraining pipelines.

Taken together, the results (Table~\ref{tab:overall_perf_final}--\ref{tab:multitask_compact}, Fig.~\ref{fig:tsne_embed}--\ref{fig:failure}) demonstrate that WLFM effectively captures both local continuity and global stratigraphic semantics. Its design choices—tokenization, stratigraphy-aware pretraining, and multi-curve multi-task learning—work synergistically to deliver state-of-the-art performance while producing interpretable and transferable embeddings.

\section{Conclusion}
\label{sec:conclusion}

We introduced WLFM, a domain-specialized foundation model for multi-curve well-log interpretation. 
By combining patch-wise tokenization, stratigraphy-aware pretraining, and multi-task fine-tuning, WLFM captures both intra-log continuity and cross-well consistency. 
Experiments demonstrate its superiority across porosity regression and lithology classification benchmarks (Table~\ref{tab:overall_perf_final}), strong scaling and few-shot adaptation (Fig.~\ref{fig:scale_plot}, Table~\ref{tab:comprehensive_comparison_single_col}), interpretable latent representations (Fig.~\ref{fig:tsne_embed}), and layer-invariant alignment across wells (Fig.~\ref{fig:tsne_layer_embedding}). 
Ablation analyses (Table~\ref{tab:ablation_tokenizer}, Fig.~\ref{fig:multicurve}, Table~\ref{tab:multitask_compact}) confirm that vector quantization, multi-curve integration, and multi-task supervision act synergistically to deliver robust generalization. 

Beyond predictive accuracy, WLFM learns a reusable geological vocabulary, exhibits emergent layer-awareness, and reconstructs masked curves with reasonable fidelity (while showing systematic offsets in shallow and ultra-deep intervals). 
Although boundary detection is not directly addressed in this work, clustering analyses reveal promising potential for future boundary-oriented applications. 
Importantly, tokenization provides a natural interface for integration with seismic and textual data, pointing toward multi-modal foundation models for holistic subsurface characterization.

Challenges remain. WLFM struggles with thinly interbedded and noisy formations (Fig.~\ref{fig:failure}), and its patch-based encoder limits modeling of long-range stratigraphic dependencies. 
Uncertainty quantification is also absent, restricting deployment in risk-sensitive applications. 
Future directions include finer-resolution or hierarchical tokenization, stratigraphy-conditioned objectives, distributed pretraining pipelines, and explicit uncertainty modeling.

Overall, WLFM establishes a scalable, interpretable, and transferable backbone for intelligent well-log interpretation. 
By bridging domain knowledge with modern foundation modeling, it provides a practical path toward next-generation geoscience AI frameworks capable of unifying logs, seismic, and textual modalities.


\section*{Acknowledgment}

This work was supported by the National Natural Science Foundation of China under Grant (No. 62273319).







\bibliographystyle{elsarticle-num}   
\bibliography{refs}                  

\end{document}